\documentclass{article}

\usepackage{arxiv}

\usepackage{amsfonts}  
\usepackage{amsmath} 
\usepackage{float}
\usepackage{longtable}
\usepackage{array}
\usepackage{makecell}
\usepackage{siunitx}
\usepackage[T1]{fontenc}    
\usepackage{hyperref}       
\usepackage{url}            
\usepackage{hyperref}
\hypersetup{
    breaklinks=true,
    colorlinks=true,
    linkcolor=blue,
    urlcolor=blue,
    citecolor=blue
}

\usepackage{booktabs}       
\usepackage{amsfonts}       
\usepackage{nicefrac}       
\usepackage{microtype}      
\usepackage{cleveref}       
\usepackage{lipsum}         
\usepackage{graphicx}
\usepackage{natbib}
\usepackage{doi}
\usepackage{parskip} 

\title{Visualization and Optimization of Continuum Robots: Integration of Lie Group Kinematics and Evolutionary Algorithm}

\date{}

\newif\ifuniqueAffiliation
\uniqueAffiliationtrue

\ifuniqueAffiliation 
\author{ \href{https://orcid.org/0009-0001-2682-138X}{\includegraphics[scale=0.06]{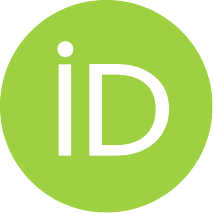}\hspace{1mm}Po-Yu Hsieh}\thanks{The demonstration video of detailed digital workflow and visualization can be found via: \url{https://youtu.be/9UWLf1npNsw}}\\
	Graduate Institute of Architecture\\
	National Yang Ming Chiao Tung University\\
	Hsinchu, Taiwan \\
	\texttt{kevinhsieh870118@arch.nycu.edu.tw} \\
	\And
	\href{https://orcid.org/0000-0002-8362-7719}{\includegraphics[scale=0.06]{orcid.pdf}\hspace{1mm}June-Hao Hou} \\
	Graduate Institute of Architecture\\
	National Yang Ming Chiao Tung University\\
	Hsinchu, Taiwan \\
	\texttt{jhou@arch.nycu.edu.tw} \\
}
\renewcommand{\headeright}{}
\renewcommand{\undertitle}{}
\renewcommand{\shorttitle}{Visualization and Optimization of Continuum Robots: Integration of Lie Group Kinematics and Evolutionary Algorithm}

\hypersetup{
pdftitle={Optimizing Modeling of Continuum Robots: Integration of Lie Group Kinematics and Evolutionary Algorithms},
pdfsubject={cs.RO},
pdfauthor={Po-Yu Hsieh, June-Hao Hou},
pdfkeywords={Robotics, Modeling and Control System, Lie Group Kinematics, Evolutionary Algorithm, Computer-Aided Design},
}
\begin{document}
\maketitle

\begin{abstract}
	Continuum robots, known for their high flexibility and adaptability, offer immense potential for applications such as medical surgery, confined-space inspections, and wearable devices. However, their non-linear elastic nature and complex kinematics present significant challenges in digital modeling and visualization. Identifying the modal shape coefficients of specific robot configuration often requires plenty of physical experiments, which is time-consuming and robot-specific. To address this issue, this research proposes a computational framework that utilizes evolutionary algorithm (EA) to simplify the coefficient identification process. Our method starts by generating datasets using Lie group kinematics and physics-based simulations, defining both ideal configurations and models to be optimized. With the deployment of EA solver, the deviations were iteratively minimized through two fitness objectives \textemdash mean square error of shape deviation (\(\text{MSE}_1\)) and tool center point (TCP) vector deviation (\(\text{MSE}_2\)) \textemdash to align the robot's backbone curve with the desired configuration. Built on the Computer-Aided Design (CAD) platform Grasshopper, this framework provides real-time visualization suitable for development of continuum robots. Results show that this integrated method achieves precise alignment and effective identification. Overall, the objective of this research aims to reduce the modeling complexity of continuum robots, enabling precise, efficient virtual simulation before robot programming and implementation.
\end{abstract}


\section{Introduction}
\subsection{Constraints of Traditional Rigid Robots}
Robotics has been gradually applied in industries as an assistive or autonomous system. Most commonly used robots in practice are based on rigid bodies, providing high payload capacity and efficiency. These rigid robots typically operate in well-defined spaces, performing repetitive, predefined tasks with exceptional accuracy \citep{trivedi2008soft}. The stiffness of current robots is designed to eliminate deformation, vibration, and other uncertain factors that may cause the deviation of movement. Therefore, the precision relied on stiffness has appreciably made them prominent in manufacturing, especially for automotive and construction industries, meeting the requirements of typical robotic applications. The fact, however, that these rigid robots often suffer from unstructured, uneven, or congested environments indicates their inability to deal with unpredictable obstacles. The absolute stiffness, while providing robust operation, limits environmental adaptability, and thus presenting a double-edged structural property.

\subsection{Significance and Challenges of Continuum Robots}
Continuum robots, with the structural integration of rigid and soft bodies, represent a different strategy compared to rigid robots. Due to their unique adaptability and high compliance in performing intricate maneuvers, they hold a significant potential for various applications in human-computer interaction (HCI), including medical surgery \citep{zhang2021design}, inspection in confined spaces \citep{kobayashi2022soft}, and wearable devices \citep{10632985}. However, their non-linear elastic nature and the inherent complexity of kinematics pose technical challenges in digital modeling and achieving effective control.

\par Traditional model-based methods often utilize Lie group kinematics \citep{9057619}, which relies on modal shape functions such as Euler curves \citep{8794238}, Legendre polynomials \citep{7339740}, or trigonometric functions \citep{8594451} to provide a mathematical framework for describing the curvature distribution of the continuum robot's central backbone. The general form of the modal shape function \(s(x)\) is given by:

\begin{equation}
   s(x) = \sum_{i=0}^{n} c_i \phi_i(x),  \quad \quad c_i = \begin{pmatrix}
    c_{ix} \\
    c_{iy} \\
    c_{iz}
    \end{pmatrix}
    \label{eqn:modalShapeFnc}
\end{equation}

where \(\phi_i(x)\) is the basis function of modal configurations, and \(c_i\) is a finite vector set of constant modal coefficients to be determined. 

\par Identifying the modal coefficients \(c_i\) usually requires an extensive amount of data collected by sensors along with numerous physical experiments, leading to practical difficulties and inefficiency in modeling and computations \citep{tariverdi2020dynamic}. The challenges are manifold. First, the non-linear behavior of continuum robots, influenced by both material properties and environmental interactions, makes it difficult to develop accurate mathematical models. Traditional approaches often fail to capture the intricate dynamics involved, especially when subjected to external forces or deformation.

\par Second, the high-dimensional nature of these robots adds complexity. Unlike rigid robots with discrete joints, continuum robots operate with continuous curvature, making their kinematic and dynamic models inherently more complex. This complexity increases the computational cost of simulations and requires advanced numerical techniques to ensure stability and accuracy during optimization. Furthermore, data collection and calibration are time-consuming tasks in practice. Real-world scenarios often introduce uncertainties, such as sensor noise, unpredictable disturbances, or material fatigue, which complicate the validation and real-time control of models \citep{qian2023feedback}. The reliance on iterative experiments for tuning parameters adds further to the time and resource demands, limiting the scalability of traditional modeling approaches.

\par Another major hurdle is the lack of standardized frameworks for continuum robots. Every design has unique structural properties—such as different material compositions, actuation methods, or degrees of freedom—which necessitates custom models and simulations. This diversity makes it difficult to generalize findings, such as simulation and calibration, across different applications or transfer solutions from one context to another, creating barriers to broader adoption \citep{modes2019calibration}.

\par To address these limitations, we propose a novel approach that establishes an end-to-end mapping between Lie group kinematics and physics-based simulations. Furthermore, facilitated by an evolutionary algorithm, this modeling approach aims to provide an effective, intuitive framework for users and developers.

\section{Proposed Method}

\subsection{Overview}

\begin{figure}[h]
	\centering
	\includegraphics[width=\linewidth]{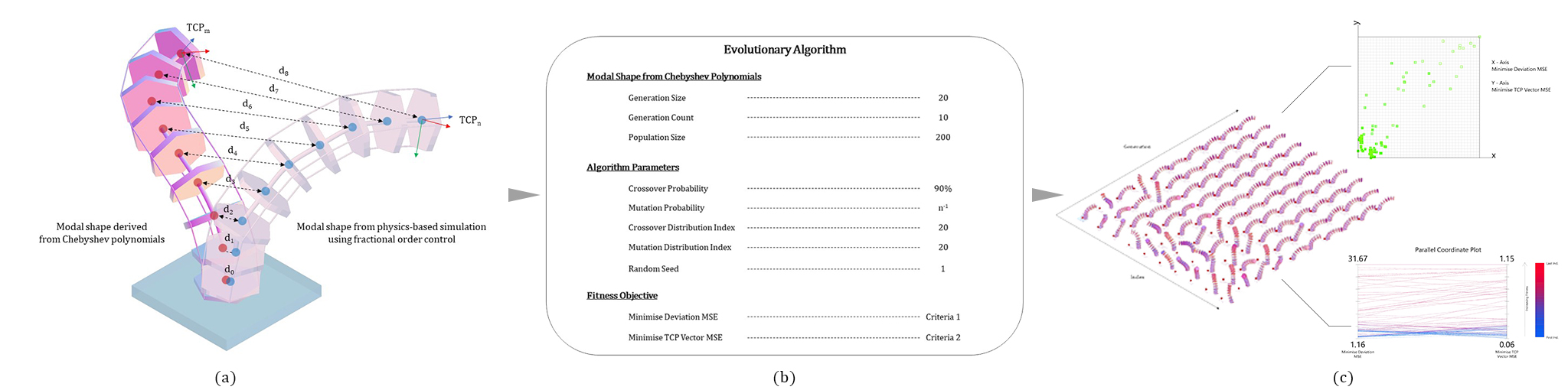}
	\caption{(a) Input dataset (b) Evolutionary algorithm (c) Generated continuum robot model.}
	\label{fig:representative}
\end{figure}

The main concept of the proposed method is to replace the traditional modal coefficients identification process with an evolutionary algorithm since it is an automated process and can provide more accuracy. To achieve, we introduce a computational framework for optimizing the continuum robot, which involves three key stages: generating datasets (Figure \ref{fig:representative}a), setting up the EA computational environment (Figure \ref{fig:representative}b), and launching the framework to achieve optimal coefficients (Figure \ref{fig:representative}c).

\begin{enumerate}
	\item \textbf{Generating Two Datasets: Ideal Configuration and Robot Model to be Optimized}
	\par In the first stage, two datasets are generated:
	\begin{itemize}
		\item Ideal Configuration: A series of ideal robot configurations are generated based on physics-based simulation. We utilize the fractional order control \cite{relano2022modeling}, which is an effective control algorithm for continuum robots. These serve as the target shapes, representing the desired outcomes for the robot's backbone curve.
		\item Robot Model for Optimization: The robot model to be optimized is derived from the modal shape function implemented through Chebyshev polynomials. This model contains initial coefficients that need to be refined to match the ideal configuration.
		\item In Figure \ref{fig:representative}a, the distance between corresponding points on the ideal configuration and the robot model is represented by $d_0$, $d_1$, \dots, $d_n$. The goal of the optimization is to minimize these distances across all points, ensuring the alignment of both shapes.
	\end{itemize}
	\item \textbf{Setting Up the EA Computational Environment}
	\par The next stage involves setting up the EA parameters and computational environment. In this research, the EA solver is configured with (Figure \ref{fig:representative}b):
	\begin{itemize}
		\item Generation Size: 20
		\item Generation Count: 10
		\item Population Size ($n$): 200
		\item Crossover Probability: 90\%
		\item Mutation Probability ($n^{-1}$): 0.5\%
		\item Fitness Objectives: Minimizing mean squared error of shape deviation and TCP Vector deviation.
	\end{itemize}
	\par These settings ensure that the algorithm explores the search space effectively and balances the two fitness objectives.
	\item \textbf{Launching the Computational Framework}
	\par In the final stage (Figure \ref{fig:representative}c), the EA framework is launched to iteratively optimize the coefficients. As the algorithm progresses, each generation refines the robot model, bringing it closer to the ideal configuration. The EA process can be visualized and robot models of each generation can be extracted, providing insight into how the coefficients evolve over time. The iterative process continues until the optimal coefficients are found, ensuring that the robot model aligns with the ideal configuration.
\end{enumerate}

\subsection{Lie Group Kinematics}

In this research, we use Lie group kinematics to model the continuous transformations of continuum robots. This method effectively handles the robot's complex deformations by representing the curvature along the backbone. Specifically, we implement the Chebyshev polynomials as the basis function \(\phi_i(x)\), which can be expressed as follows:

\begin{equation}
  \phi_0 = 1, \quad \phi_1(x) = x, \quad
  \phi_i(x) = 2xT_{i-1}(x) - T_{i-2}(x), \quad i = 2, 3, \ldots
  \label{eqn:chebyshev}
\end{equation}

where \(\phi_i(x)\) is the \(i^\text{th}\) degree Chebyshev polynomial, and $T(s)$ is the backbone transformation matrices. Based on the study \cite{10013748}, the Chebyshev polynomial is suitable in this context since it ensures simplified computations of the admissible robot workspace.
These transformations describe the relationship between points along the backbone at different lengths $s$ along the curve. The central backbone is a continuous structure, and the transformation along its length captures both translation and rotation behaviors essential for the robot's flexible movement. This transformation matrices of backbone curve can be considered as:

\begin{itemize}
	\item $T(0)$: The transformation matrix at the origin of the backbone. This matrix represents the initial pose of the robot's base in 3D space, defining the initial orientation in terms of x, y, and z axes.
\end{itemize}

\begin{itemize}
	\item $T(L)$: This matrix represents the transformation at the end of the backbone, at length $L$. It shows the cumulative effect of all transformations along the backbone from the origin to the tip.
\end{itemize}

\begin{itemize}
	\item $T(s)$: Any point along the backbone at length $s$ can be expressed by an intermediate transformation matrix $T(s)$, which captures the local curvature and orientation at that point.
\end{itemize}

Building upon the transformation matrices, the detailed computational pipeline we used can be summarized as:

\begin{enumerate}
    \item \textbf{Recursive Relation between Segments:}
    \begin{itemize}
        \item The transformation matrix at any point \(T(s)\) can be expressed as:
        \[
        T(s) = T(s - 1) \times u(s)
        \]
        \item Here, \(T(s - 1)\) is the transformation matrix from the previous segment, and \(u(s)\) represents the local curvature distribution between the two adjacent segments \(s - 1\) and \(s\).
    \end{itemize}
    
    \item \textbf{Configuration of the Robot's Pose:}
    \begin{itemize}
        \item The state of the robot at any point can be represented by a combination of these transformations along the backbone. Thus, the configuration of the entire robot can be described as the distribution of curvature along the backbone:
        \[
			\mathcal{C} \propto \left\{ T(s) \right\}_{s=0}^{L}
        \]
		\par where \(\mathcal{C}\) denotes the robot configuration space.
        \item This formulation allows for flexible modeling, where both position and orientation are derived directly from the curvature along the robot's backbone.
    \end{itemize}
    
    \item \textbf{Dynamic Motion Representation:}
    \begin{itemize}
        \item The robot's movement is captured as incremental transformations along the original backbone curve. Each transformation between consecutive points describes a linear translation or rotational adjustment.
        \item For instance, at any given time step, the robot's updated state is:
        \[
        T'(s) = T(s) \times n(s)
        \]
        \item Here, \(n(s)\) represents a small adjustment of both rotation and translation applied to the backbone curve at segment \(s\).
        \item Eventually, by interpolating each distributed point and corresponding plane normal as perpendicular frames, the backbone curve can be visualized.
    \end{itemize}
\end{enumerate}

\begin{figure}[h]
	\centering
	\includegraphics[width=.75\linewidth]{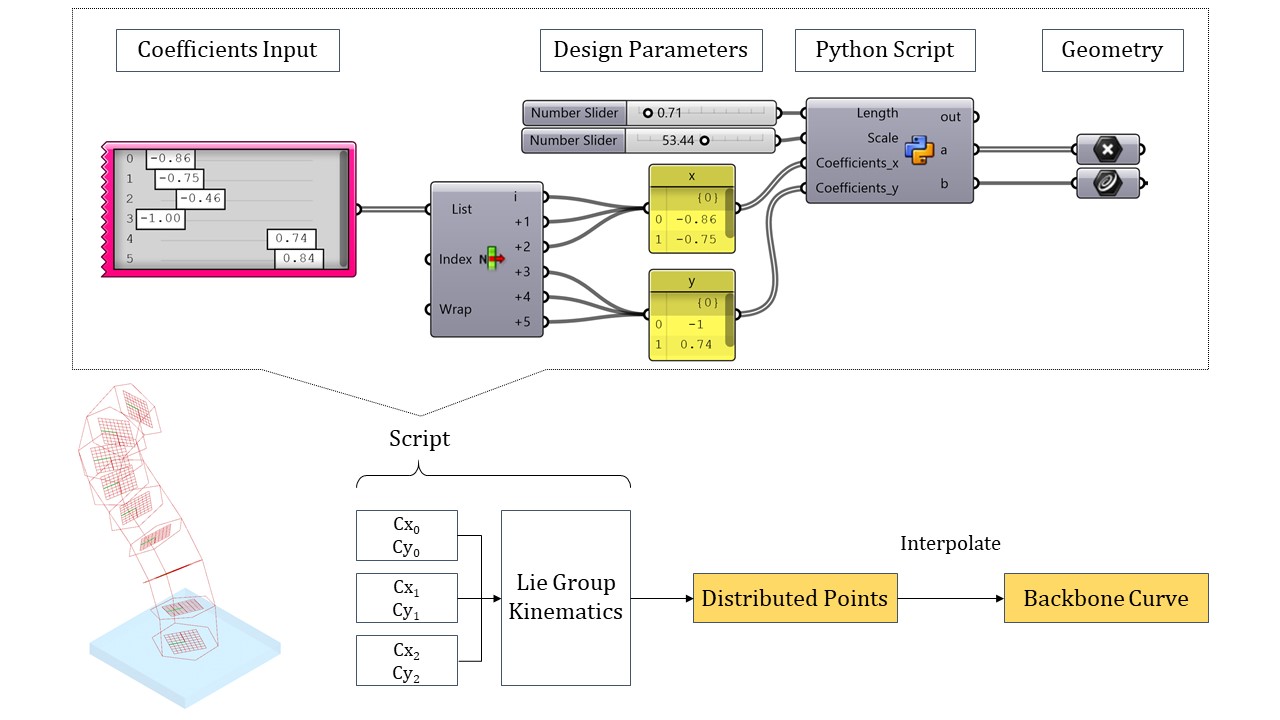}
	\caption{Computational framework based on Grasshopper.}
	\label{fig:lieGroupEvo}
\end{figure}

Following this motion representation idea, we established a computational framework in Grasshopper, a CAD platform integrated with parametric design tools, as visualized in Figure \ref{fig:lieGroupEvo}. The following describes the step-by-step implementation of this transformation algorithm on Grasshopper:

\begin{itemize}
	\item Coefficients Input: The sliders allow for dynamic changes in control parameters $c_x$ and $c_y$. In this research, we use three sets of inputs for each $c_x$ and $c_y$, representing Chebyshev polynomials of degree 3.
\end{itemize}

\begin{itemize}
	\item Design Parameters: Parameters for dimensions (overall length and scale) are fed into the Python script as well.
\end{itemize}

\begin{itemize}
	\item Python Script: It calculates the backbone curve's shape based on Lie group kinematics.
\end{itemize}

\begin{itemize}
	\item Geometry: The resulting backbone curve is rendered by interpolating distributed points.
\end{itemize}

The Python script implemented in Grasshopper is as follows:

\begin{verbatim}
import Rhino.Geometry as rg

# Define Chebyshev polynomials
def chebyshev_polynomials(n, x):
    T = [0] * (n + 1)
    T[0] = 1
    if n > 0:
        T[1] = x
    for i in range(2, n + 1):
        T[i] = 2 * x * T[i - 1] - T[i - 2]
    return T
    
# Define curvature distribution    
def curvature_distribution(coeffs, s, L):
    u = []
    for i in range(len(s)):
        x = 2 * s[i] / L - 1
        T = chebyshev_polynomials(len(coeffs) - 1, x)
        u.append(sum(c * t for c, t in zip(coeffs, T)))
    return u

# Define twist transformation for each prep frame
def integrate_twist(u_x, u_y, s):
    T = rg.Transform.Identity
    points = [rg.Point3d(0, 0, 0)]
    
    for i in range(1, len(s)):
        ds = s[i] - s[i - 1]
        theta_x = u_x[i] * ds
        theta_y = u_y[i] * ds
        
        Rx = rg.Transform.Rotation(theta_x, rg.Vector3d(1, 0, 0), rg.Point3d.Origin)
        Ry = rg.Transform.Rotation(theta_y, rg.Vector3d(0, 1, 0), rg.Point3d.Origin)
        
        T = T * Rx * Ry
        
        p = rg.Point3d(0, 0, s[i])
        p.Transform(T)
        p = rg.Point3d(p.X * scale, p.Y * scale, p.Z * scale)  # Scale the point
        points.append(p)
    
    return points
    
# Parameters
L = Length  # Total length of the backbone
s = [i / 100.0 * L for i in range(101)]  # Discretize the length
coeffs_x = Coefficients_x  # coefficients input
coeffs_y = Coefficients_y  # coefficients input
scale = Scale

# Compute curvature
u_x = curvature_distribution(coeffs_x, s, L)
u_y = curvature_distribution(coeffs_y, s, L)
#u_y = [0] * len(s)  # Assume no curvature in the y-direction for simplicity

# Integrate to find the transformation matrices
points = integrate_twist(u_x, u_y, s)

# Create an interpolated curve through the points
curve = rg.Curve.CreateInterpolatedCurve(points, 3)  # Degree 3 interpolated curve

# Output to Grasshopper
a = points  # Output the points to Grasshopper
b = curve   # Output the interpolated curve to Grasshopper
\end{verbatim}

\begin{figure}[H]
    \centering
    \begin{minipage}[t]{.33\textwidth}
        \centering
		\includegraphics[width=\linewidth]{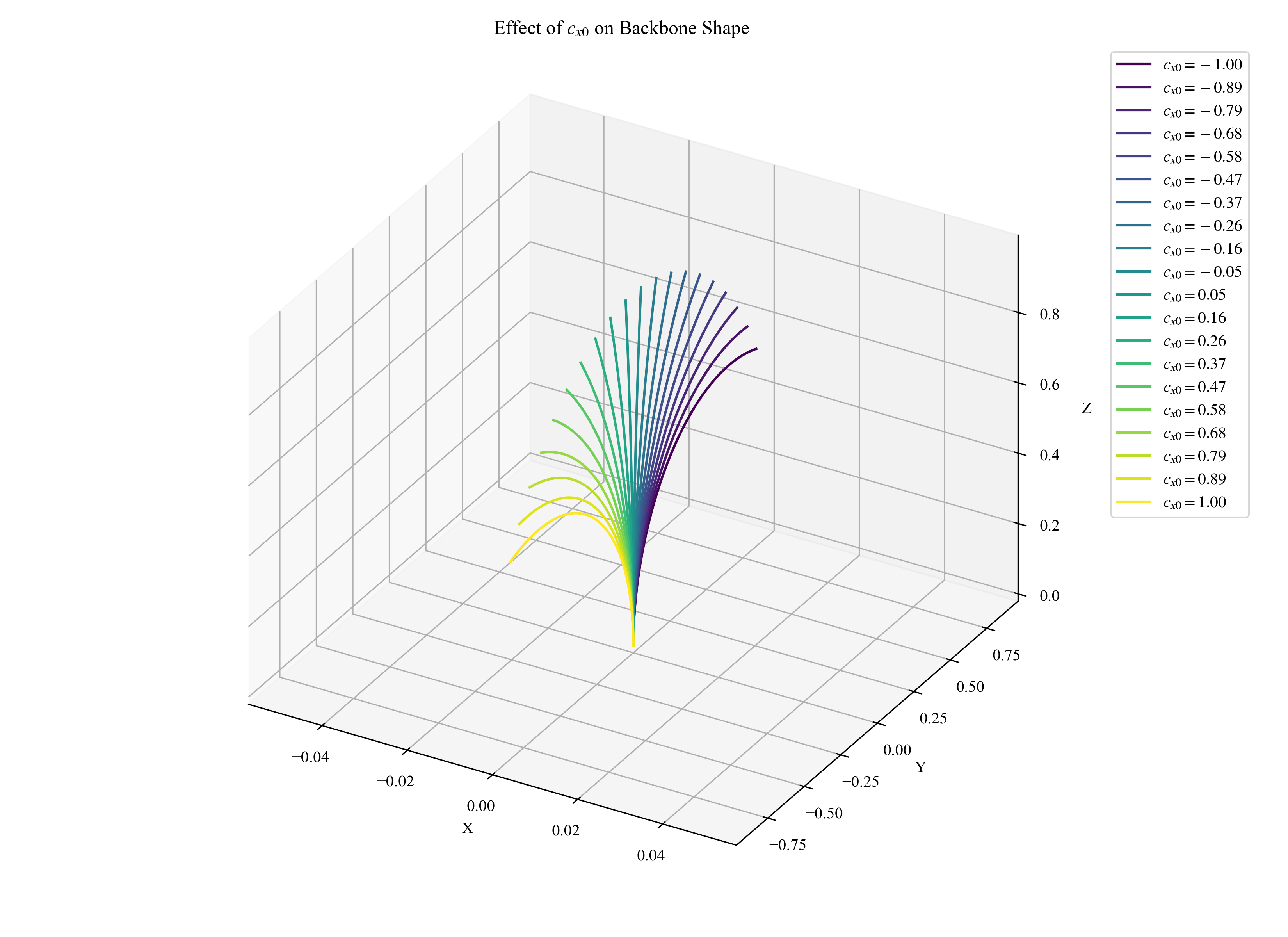}
 		 \caption{C$x_0$.}
  		\label{fig:cx0}
    \end{minipage}
    \hfill
    \begin{minipage}[t]{.33\textwidth}
        \centering
        \includegraphics[width=\linewidth]{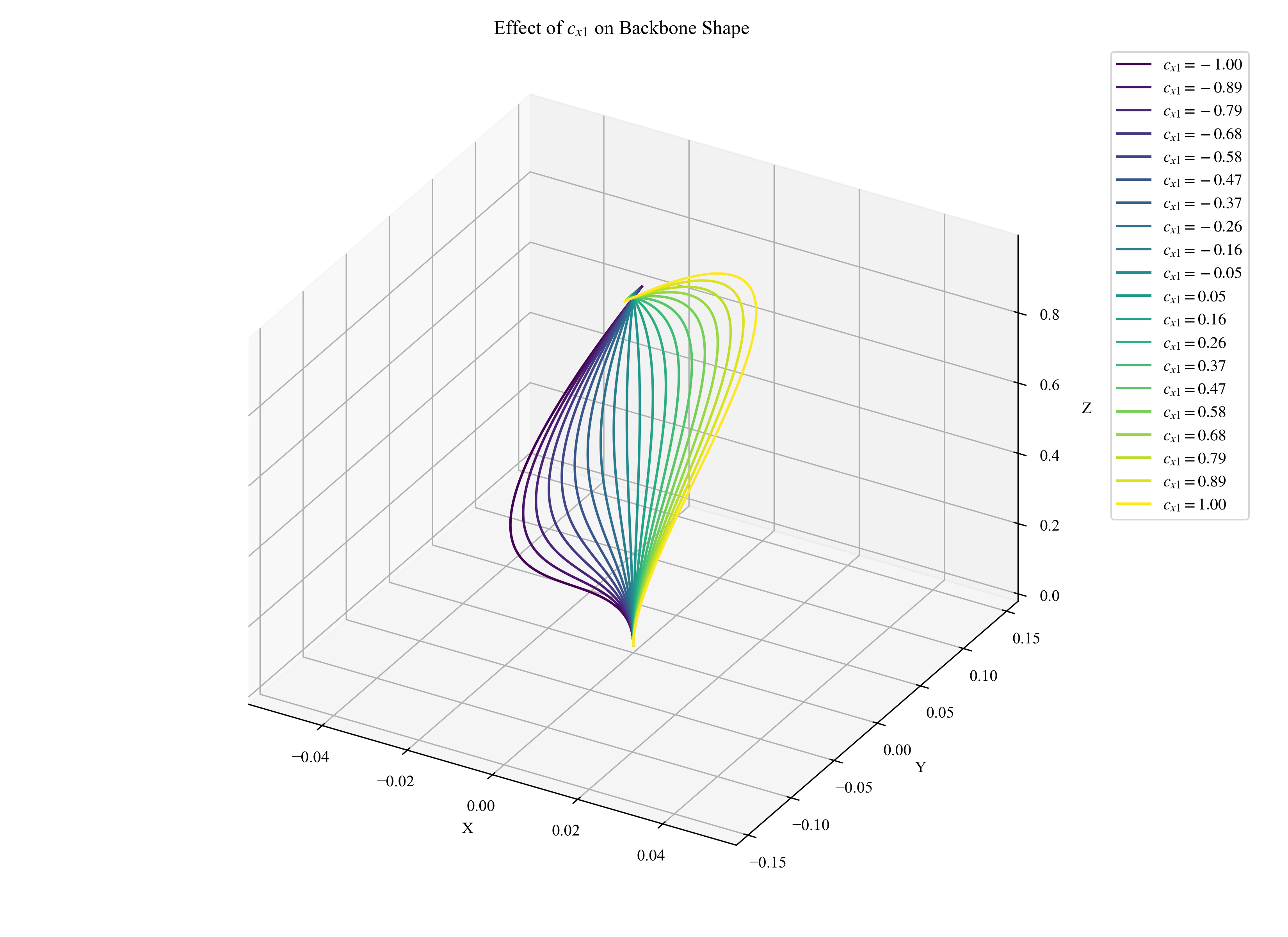}
 		 \caption{C$x_1$.}
  		\label{fig:cx1}
    \end{minipage}
	\hfill
    \begin{minipage}[t]{.33\textwidth}
        \centering
        \includegraphics[width=\linewidth]{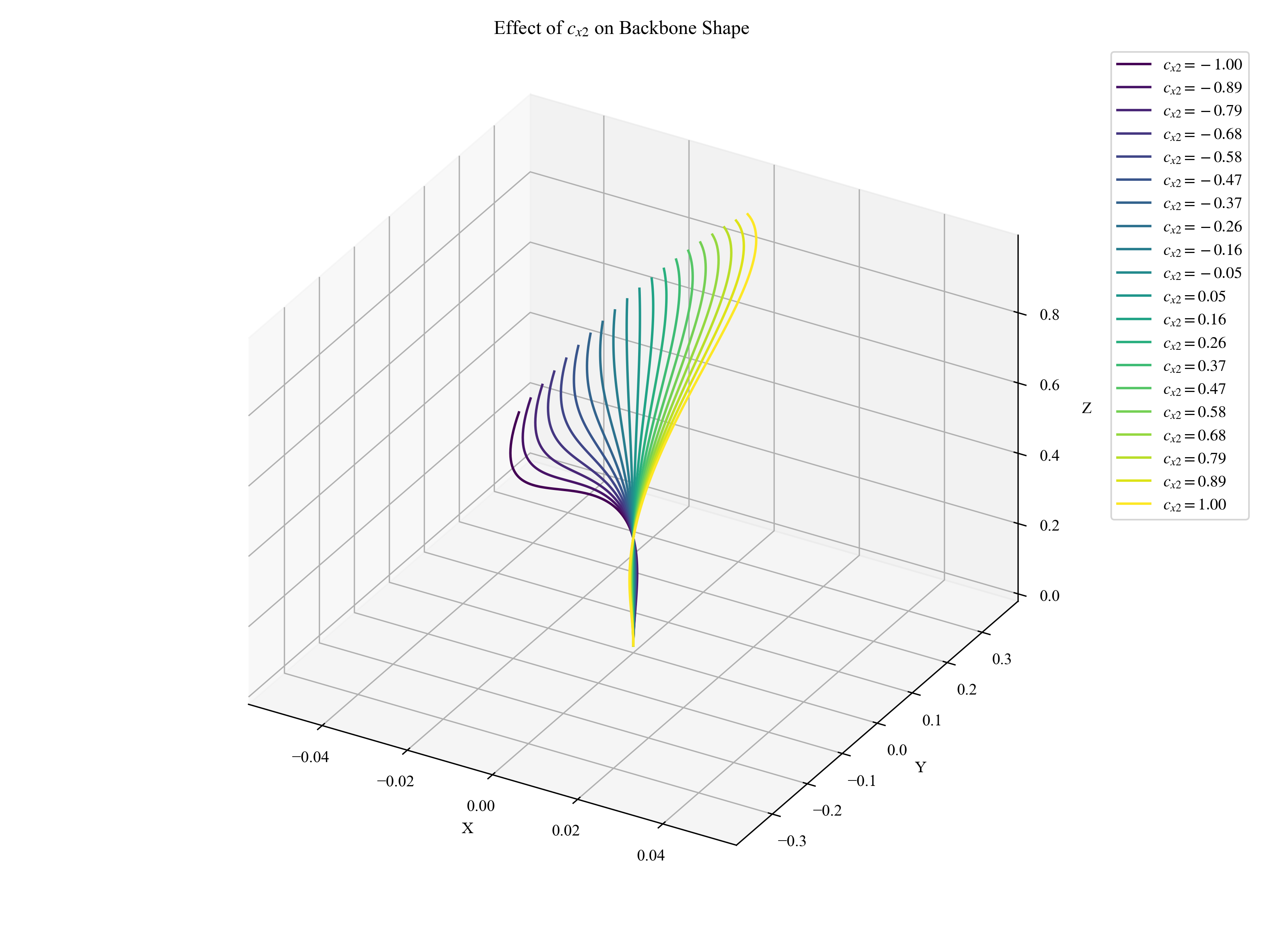}
 		 \caption{C$x_2$.}
  		\label{fig:cx2}
    \end{minipage}
\end{figure}

\begin{figure}[H]
    \centering
    \begin{minipage}[t]{.33\textwidth}
        \centering
		\includegraphics[width=\linewidth]{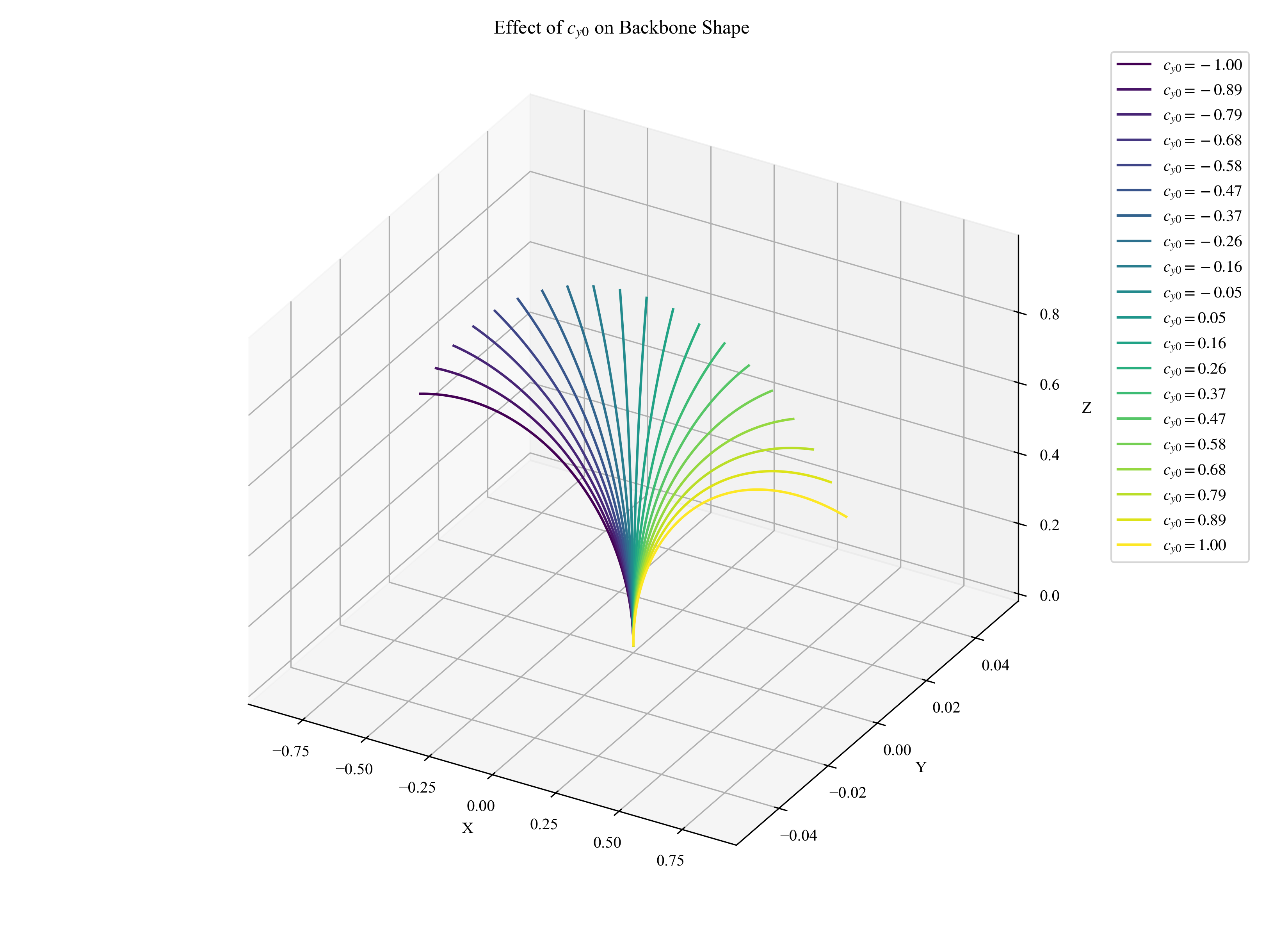}
 		 \caption{C$y_0$.}
  		\label{fig:cy0}
    \end{minipage}
    \hfill
    \begin{minipage}[t]{.33\textwidth}
        \centering
        \includegraphics[width=\linewidth]{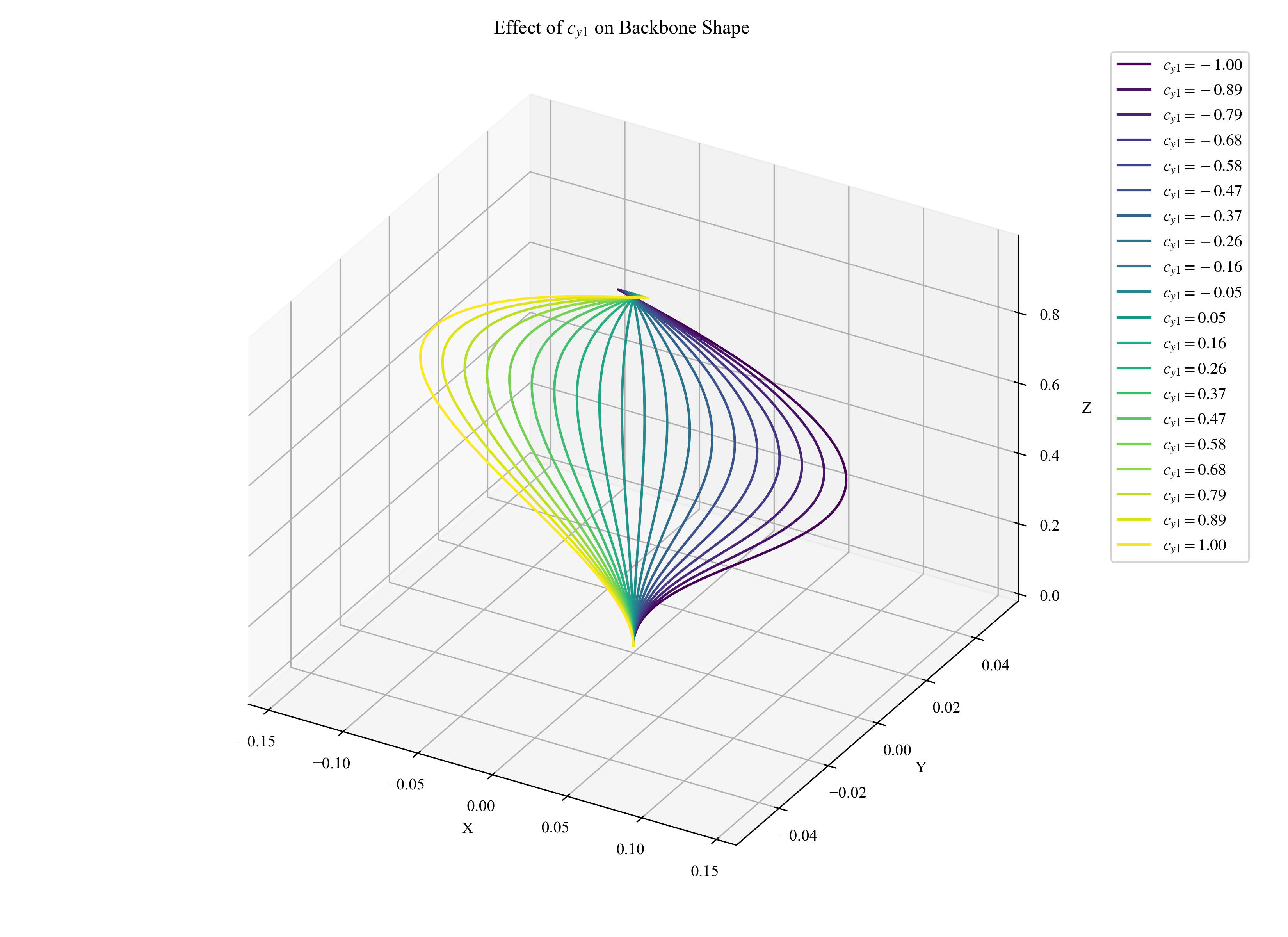}
 		 \caption{C$y_1$.}
  		\label{fig:cy1}
    \end{minipage}
	\hfill
    \begin{minipage}[t]{.33\textwidth}
        \centering
        \includegraphics[width=\linewidth]{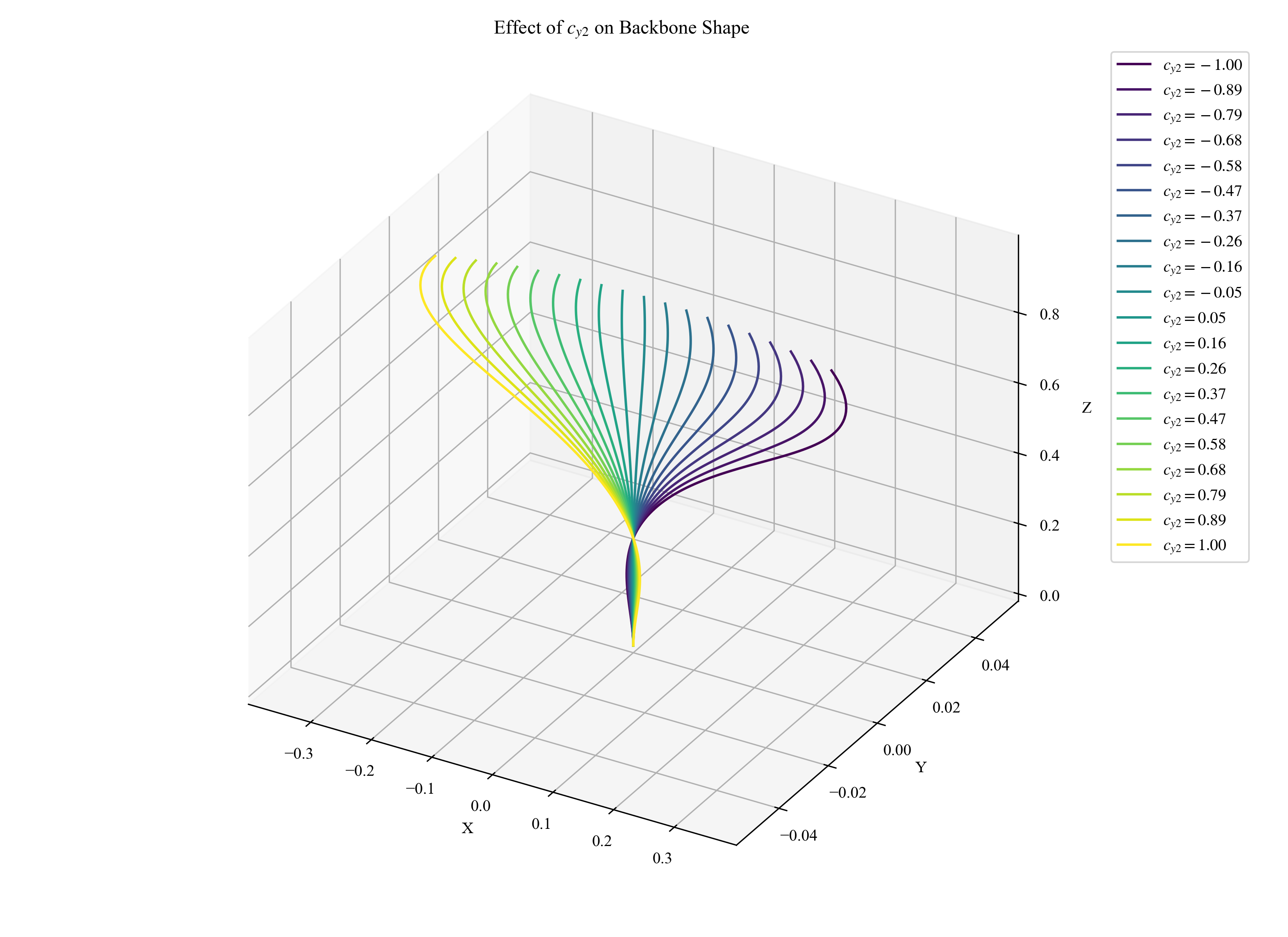}
 		 \caption{C$y_2$.}
  		\label{fig:cy2}
    \end{minipage}
\end{figure}

\par Based on the proposed framework, we can control the robot configuration by adjusting the coefficients. Thus, we tested the impact on backbone shape resulted from different conefficient inputs, which can be summarized as:

\begin{itemize}
	\item $T_0(x)$ (C$x_0$) and $T_0(y)$ (C$y_0$): This term sets the initial translational movements along the x- and y-axes. It's responsible for defining the starting position and orientation of the robot's backbone. Since it's constant, it doesn't introduce any curvature or bending. Instead, it establishes the base configuration, meaning it dictates where the backbone begins in space (Figure \ref{fig:cx0} and \ref{fig:cy0}).
\end{itemize}

\begin{itemize}
	\item $T_1(x)$ (C$x_1$) and $T_1(y)$ (C$y_1$): This term introduces simple bending. It accounts for linear changes in curvature, which means the robot will bend gradually from its initial configuration. It essentially governs the basic deformations of the robot. As you adjust coefficient C$x_1$ or C$y_1$, the robot's backbone will smoothly bend in one direction (Figure \ref{fig:cx1} and \ref{fig:cy1}).
\end{itemize}

\begin{itemize}
	\item $T_2(x)$ (C$x_2$) and $T_2(y)$ (C$y_2$): This term introduces higher-order bending patterns. Unlike the linear term, this quadratic term allows for complex deformations. As this term grows in influence, the robot can exhibit more intricate shapes, which may involve multiple bends or more sophisticated curvatures along its backbone (Figure \ref{fig:cx2} and \ref{fig:cy2}).
\end{itemize}

\subsection{Evolutionary Algorithm}

To identify an appropriate coefficients combination, we integrate the evolutionary algorithm into our framework using the Wallacei plugin within Grasshopper. The method of using the evolutionary algorithm involves the following steps:

\begin{enumerate}
	\item \textbf{Generating Ideal Datasets}: We first generate a series of ideal robot configurations using fractional order control and physics-based simulations on the Grasshopper platform. These configurations represent the target shapes for the robot's backbone and modal shape function \(s(x)\), serving as benchmarks for optimization.
	\item \textbf{Connecting Ideal and Generated Models}: The ideal configurations are then connected to the models generated from the previous section's Lie Group Kinematics framework. This connection allows the evolutionary algorithm to compare the generated models against the ideal shapes.
	\item \textbf{Setting Fitness Objectives}: We define two fitness objectives based on the deviations between the generated models and the ideal configurations: Minimize deviation mean squared error (\(\text{MSE}_1\)) and Minimize TCP vector mean squared error(\(\text{MSE}_2\)), which can be summarized as follows:
	\begin{itemize}
		\item Minimize deviation mean squared error (\(\text{MSE}_1\))
	  \end{itemize}
	  
	  \begin{equation}
		\text{MSE}_1(u) = \sum_{i=0}^{n} (d_i)^2 = \frac{1}{n} \sum_{i=0}^{n} (u_i - \hat{u}_i)^2
	  \end{equation}
	  
	  where $n$ is the division count of the central backbone curve (with $n = 8$ in this project), $u_i$ is the position of the $i$th division point on the curve, $\hat{}$ operator denotes the ideal shape, and $d_i$ is the distance between the $i$th points of the generated and ideal curve.
	  
	  \begin{itemize}
		\item Minimize TCP vector mean squared error (\(\text{MSE}_2\))
	  \end{itemize}
	  
	  \begin{equation}
		\text{MSE}_2(v) = \frac{1}{n} \sum_{i=1}^{n} \left\lvert v - \hat{v}\right\rvert^2
	  \end{equation}

	  where $v$ is the tool center point (TCP$_m$) vector of the curve derived from the modal shape function, and $\hat{v}$ indicates the TCP (TCP$_n$) vector from the ideal shape.

	  \item \textbf{Iterative Minimization through Generations}: With the fitness objectives set, the evolutionary algorithm iteratively minimizes the deviations across generations. Each generation refines the coefficients based on the fitness values, gradually converging toward optimal solutions.
	  \item \textbf{Parameter Settings and Execution}: We run the algorithm with 10 generations and 20 individuals per generation. At each iteration, the algorithm evaluates the fitness of all individuals and retains the best solutions to evolve toward optimal configurations. The refined coefficients after several generations yield a model that closely matches the ideal configurations.
\end{enumerate}

\section{Results}

\begin{figure}[H]
    \centering
    \begin{minipage}[t]{.49\textwidth}
        \centering
        \includegraphics[width=\linewidth]{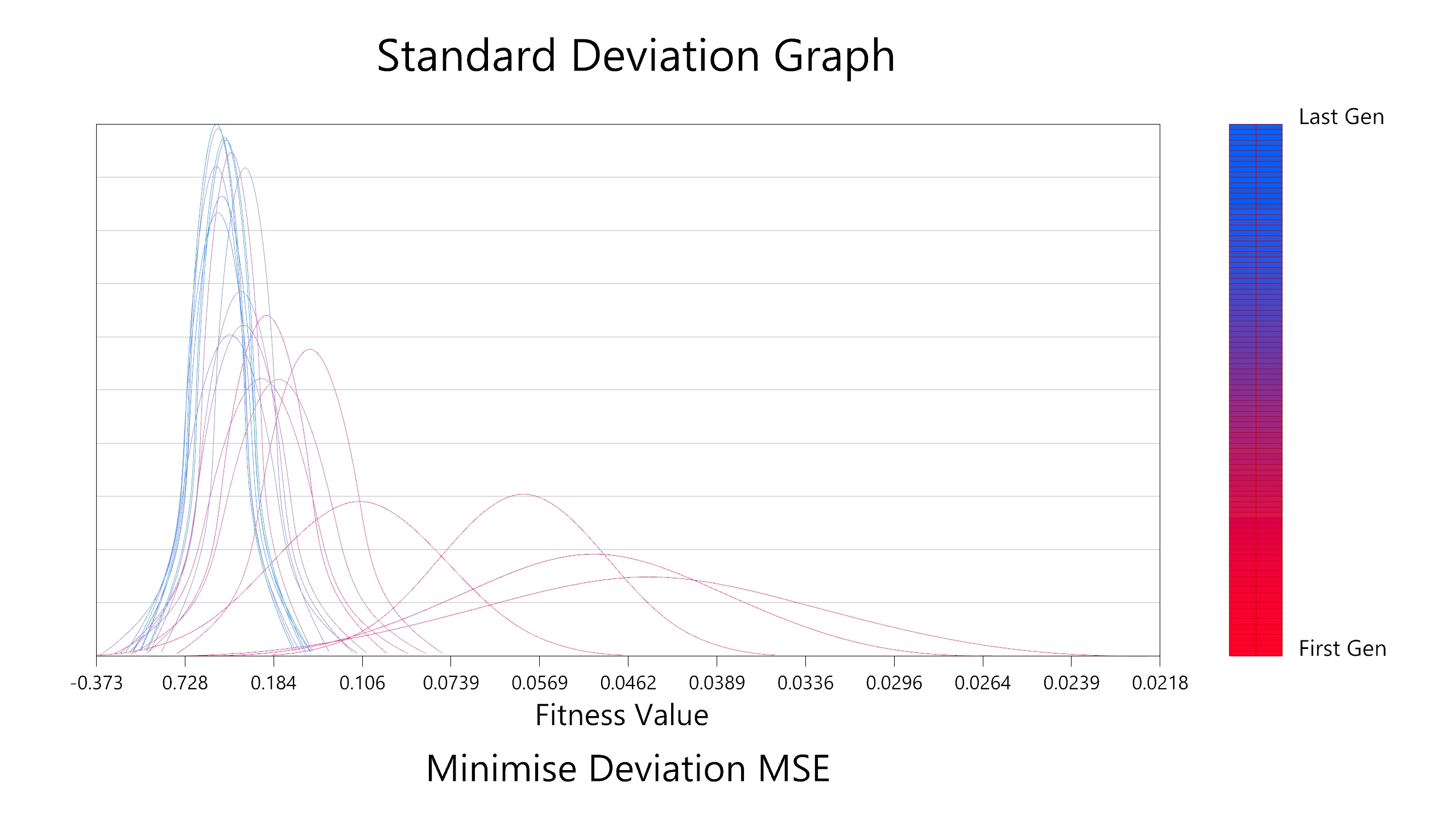}
  		\caption{Standard deviation with MSE$_1$.}
  		\label{fig:crit1_standardDeviation}
    \end{minipage}
    \hfill
    \begin{minipage}[t]{.49\textwidth}
        \centering
        \includegraphics[width=\linewidth]{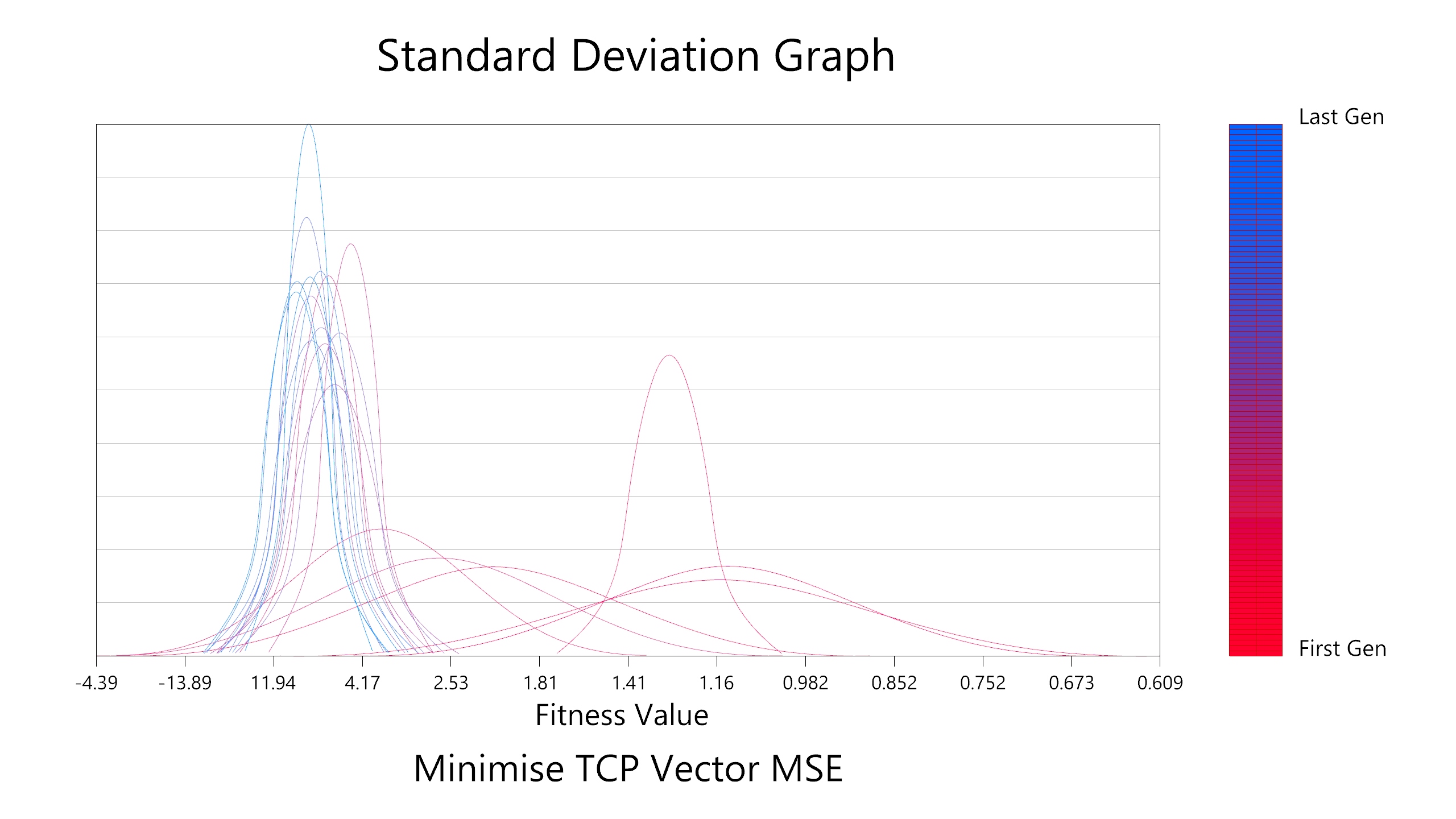}
  		\caption{Standard deviation with MSE$_2$.}
  		\label{fig:crit2_standardDeviation}
    \end{minipage}
\end{figure}

\begin{figure}[H]
    \centering
    \begin{minipage}[t]{.49\textwidth}
        \centering
		\includegraphics[width=\linewidth]{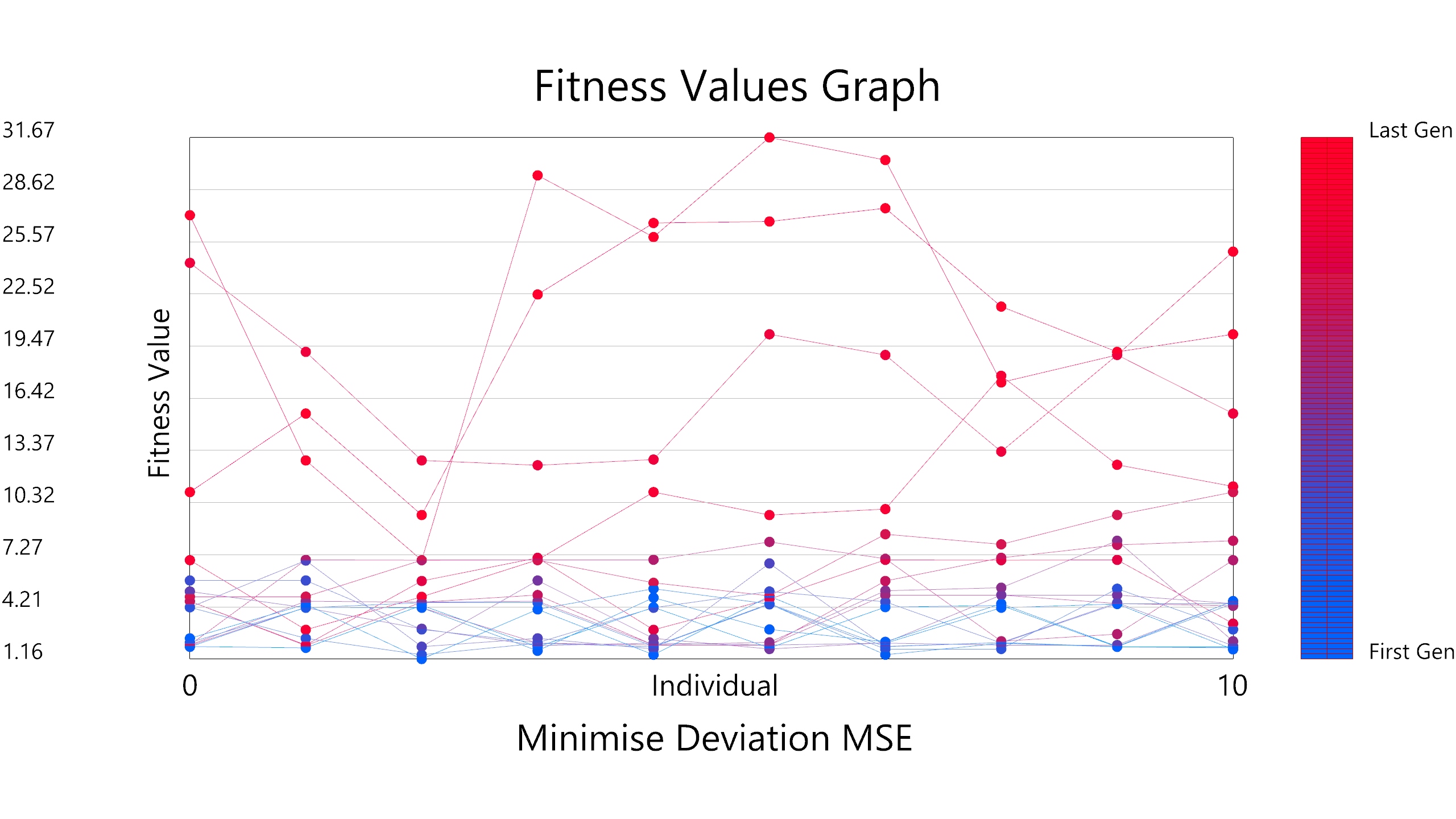}
 		 \caption{Fitness values with MSE$_1$.}
  		\label{fig:crit1_fitnessValues}
    \end{minipage}
    \hfill
    \begin{minipage}[t]{.49\textwidth}
        \centering
        \includegraphics[width=\linewidth]{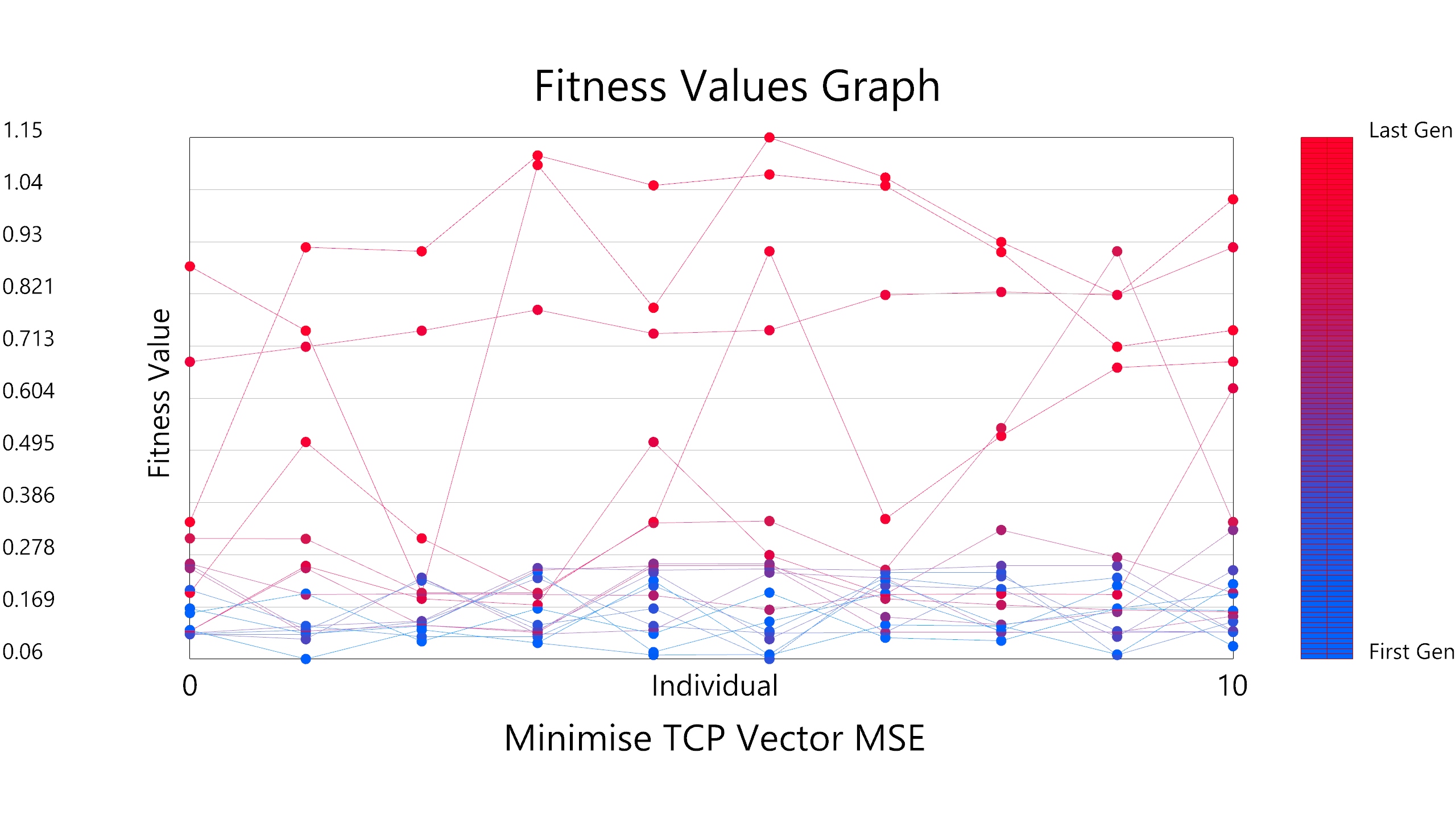}
 		 \caption{Fitness values with MSE$_2$.}
  		\label{fig:crit2_fitnessValues}
    \end{minipage}
\end{figure}

The optimization process, driven by evolutionary algorithms, demonstrates steady improvement across multiple generations. Figure \ref{fig:crit1_standardDeviation} and \ref{fig:crit2_standardDeviation} track the evolution of the standard deviation factor across generations during the optimization process. Initially, the curve is flat, reflecting significant variability in the dataset. However, as generations progress, it becomes narrower, indicating that the population's fitness values are converging towards a more uniform distribution. 

\par From an algorithmic standpoint, the reduction in standard deviation demonstrates the efficiency of the evolutionary algorithm in finding optimal or near-optimal solutions. The graphs also reflect the success of multi-objective optimization by showing that the algorithm can minimize both deviation error and TCP error simultaneously.

\par For continuum robots, where maintaining precise configurations is essential, minimizing deviation in both shape and TCP error directly impacts the robot's accuracy in performing tasks. This is especially important for applications, where even minor deviations can lead to significant errors during operation.

\par Figure \ref{fig:crit1_fitnessValues} and \ref{fig:crit2_fitnessValues} illustrate the fitness value progression of individuals across generations, measured with MSE$_1$ and MSE$_2$. These graphs help visualize the evolutionary trajectory of each individual and the effectiveness of the fitness function in guiding the population toward better solutions. In both graphs, the fitness values (represented by dots) show a downward trend as the algorithm progresses from Generation 0 to Generation 9, with red indicating earlier generations and blue representing later generations. This trend reflects the improvement in individual performance through selection and mutation processes. As the population evolves, the gap between high- and low-fitness individuals narrows, indicating that the algorithm successfully minimizes outliers and stabilizes the population toward a uniformly optimal state.

\par The improvement in fitness values demonstrates the capability of evolutionary algorithms to explore and exploit the search space effectively. The graphs highlight the balance between exploration (diversity in the early stages) and exploitation (convergence in later stages), which is critical for avoiding local optima in optimization problems.

\begin{figure}[H]
    \centering
    \begin{minipage}[t]{.49\textwidth}
        \centering
        \includegraphics[width=\linewidth]{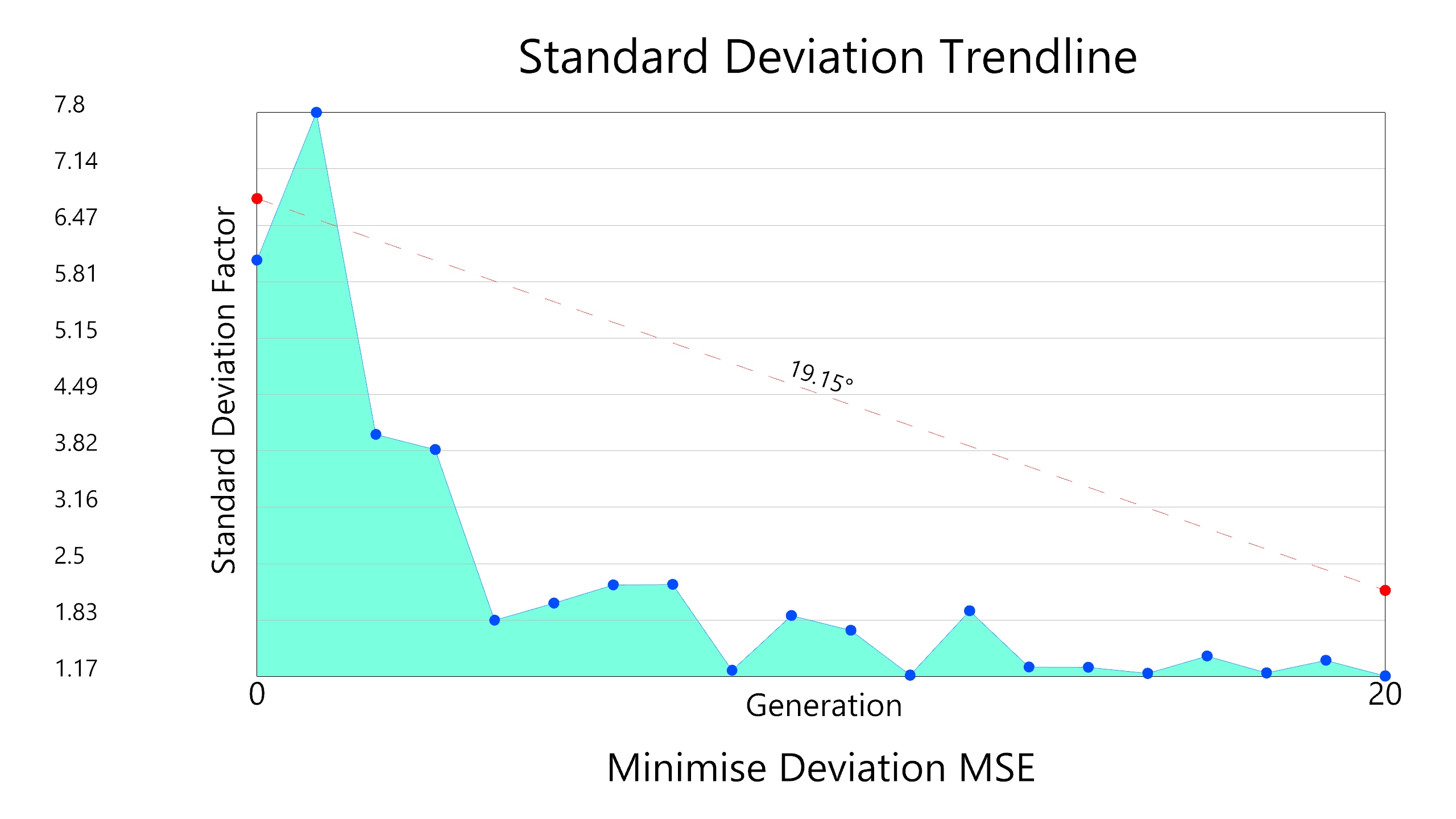}
  		\caption{Standard deviation trendline with MSE$_1$.}
  		\label{fig:crit1_standardDeviationTrendline}
    \end{minipage}
    \hfill
    \begin{minipage}[t]{.49\textwidth}
        \centering
        \includegraphics[width=\linewidth]{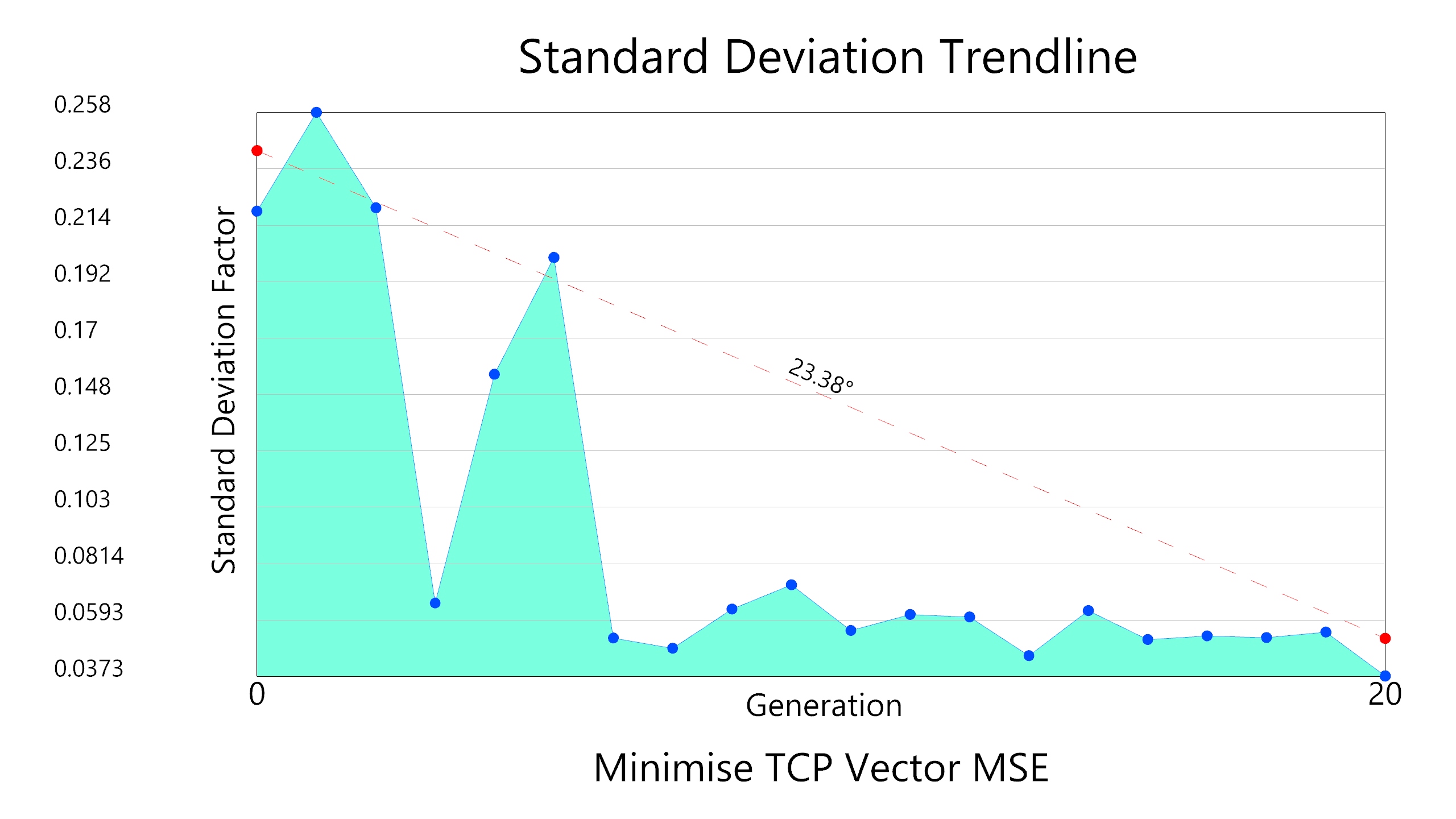}
 		 \caption{Standard deviation trendline with MSE$_2$.}
  		\label{fig:crit2_standardDeviationTrendline}
    \end{minipage}
\end{figure}

\begin{figure}[H]
    \centering
    \begin{minipage}[t]{.49\textwidth}
        \centering
        \includegraphics[width=\linewidth]{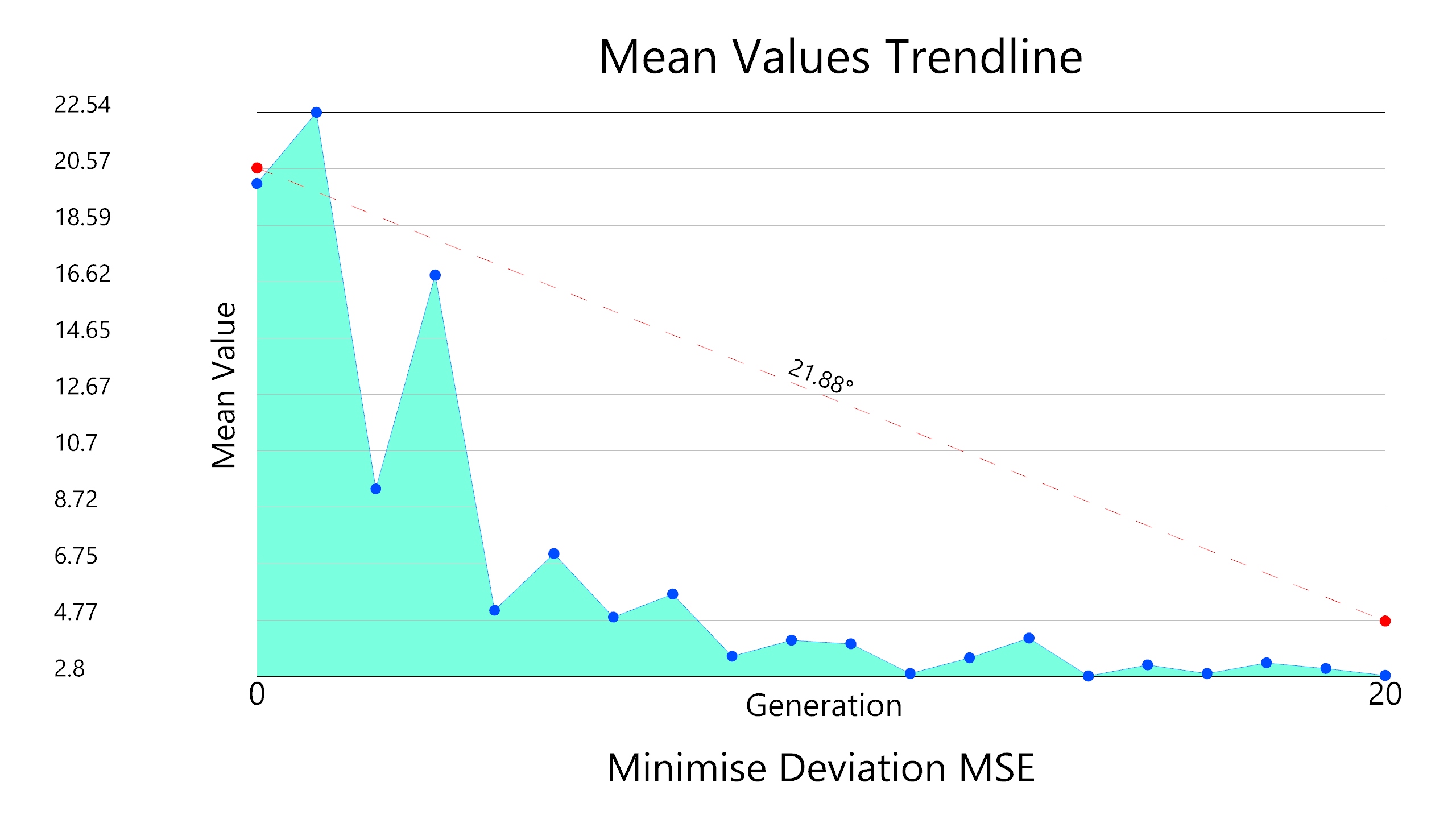}
  		\caption{Mean values trendline with MSE$_1$.}
  		\label{fig:crit1_meanValuesTrendline}
    \end{minipage}
    \hfill
    \begin{minipage}[t]{.49\textwidth}
        \centering
        \includegraphics[width=\linewidth]{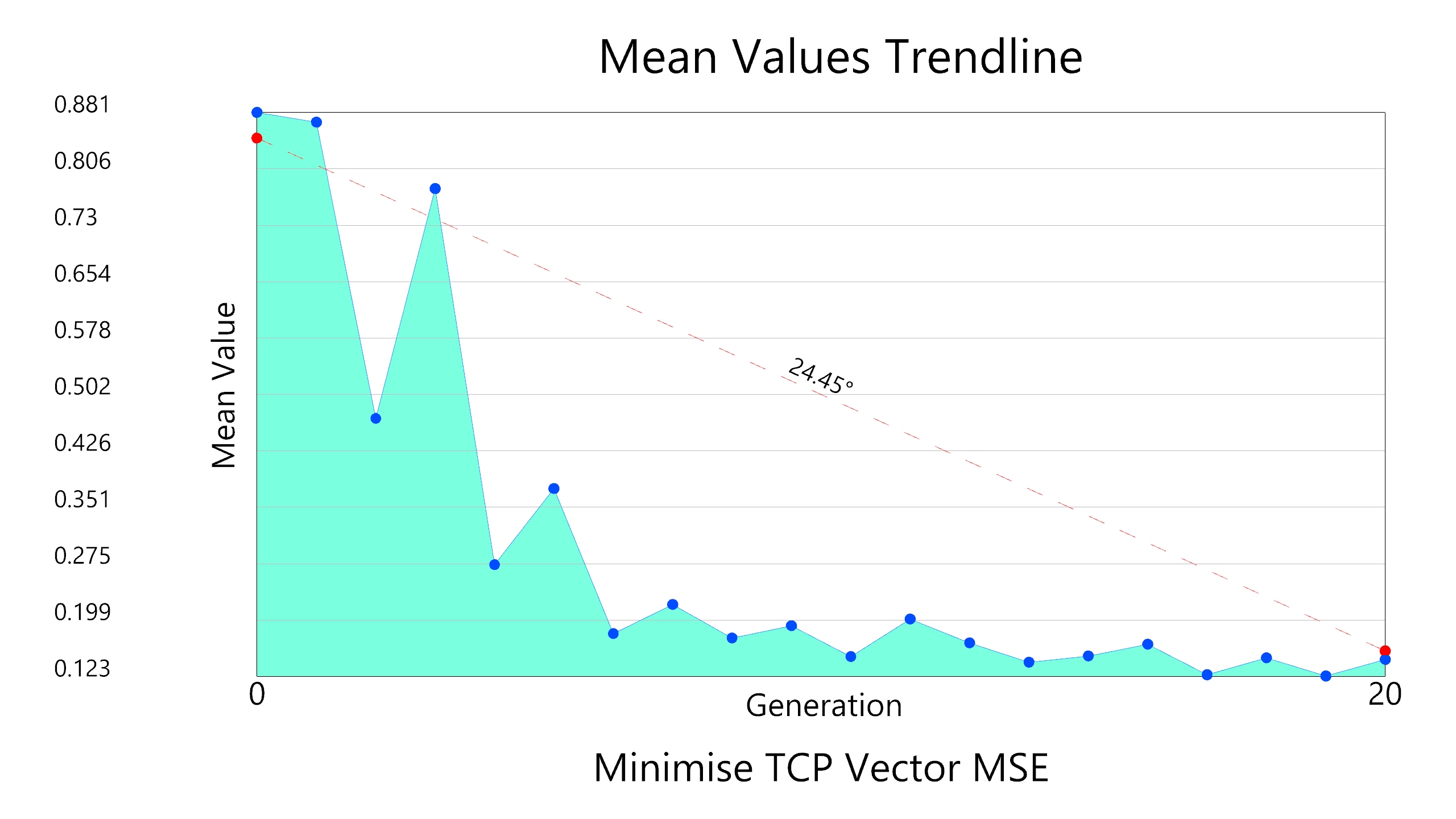}
 		 \caption{Mean values trendline with MSE$_2$.}
  		\label{fig:crit2_meanValuesTrendline}
    \end{minipage}
\end{figure}

\par In continuum robotics, achieving precise TCP control is crucial for real-world applications. The steady improvement in fitness values with MSE$_2$ shows that the robot model's control accuracy improves over time, enhancing its performance for tasks requiring delicate operations, such as remote inspections or medical interventions. Additionally, the minimized shape deviation reflected by MSE$_1$ ensures that the robot maintains smooth and predictable movements, which is vital for applications involving continuous contact with the environment.

\par Figure \ref{fig:crit1_standardDeviationTrendline} and \ref{fig:crit2_standardDeviationTrendline} present the timeline of standard deviation, while figure \ref{fig:crit1_meanValuesTrendline} and \ref{fig:crit2_meanValuesTrendline} illustrate the mean fitness values over successive generations. The initial steep decline in mean values suggests rapid early optimization, with the algorithm quickly identifying and favoring more fit individuals. As the evolutionary algorithm operates, the rate of improvement slows, and the mean values level off, indicating that the population is approaching an optimal solution. The trendlines quantify this improvement rate, underscoring the effectiveness of the optimization process in enhancing the overall fitness of the population.

\par From a robotics perspective, the results validate the algorithm's capacity to model the continuum robot with increasing accuracy across generations. The minimized standard deviation at later generations indicates that the generated configurations are becoming more consistent, crucial for tasks such as precise assembly, inspection, or surgery, where stability and repeatability are key.

\subsection{Modeling and Visualization}

\begin{figure}[htbp]
    \centering
    \begin{minipage}[t]{.3\textwidth}
        \centering
		\includegraphics[width=\linewidth]{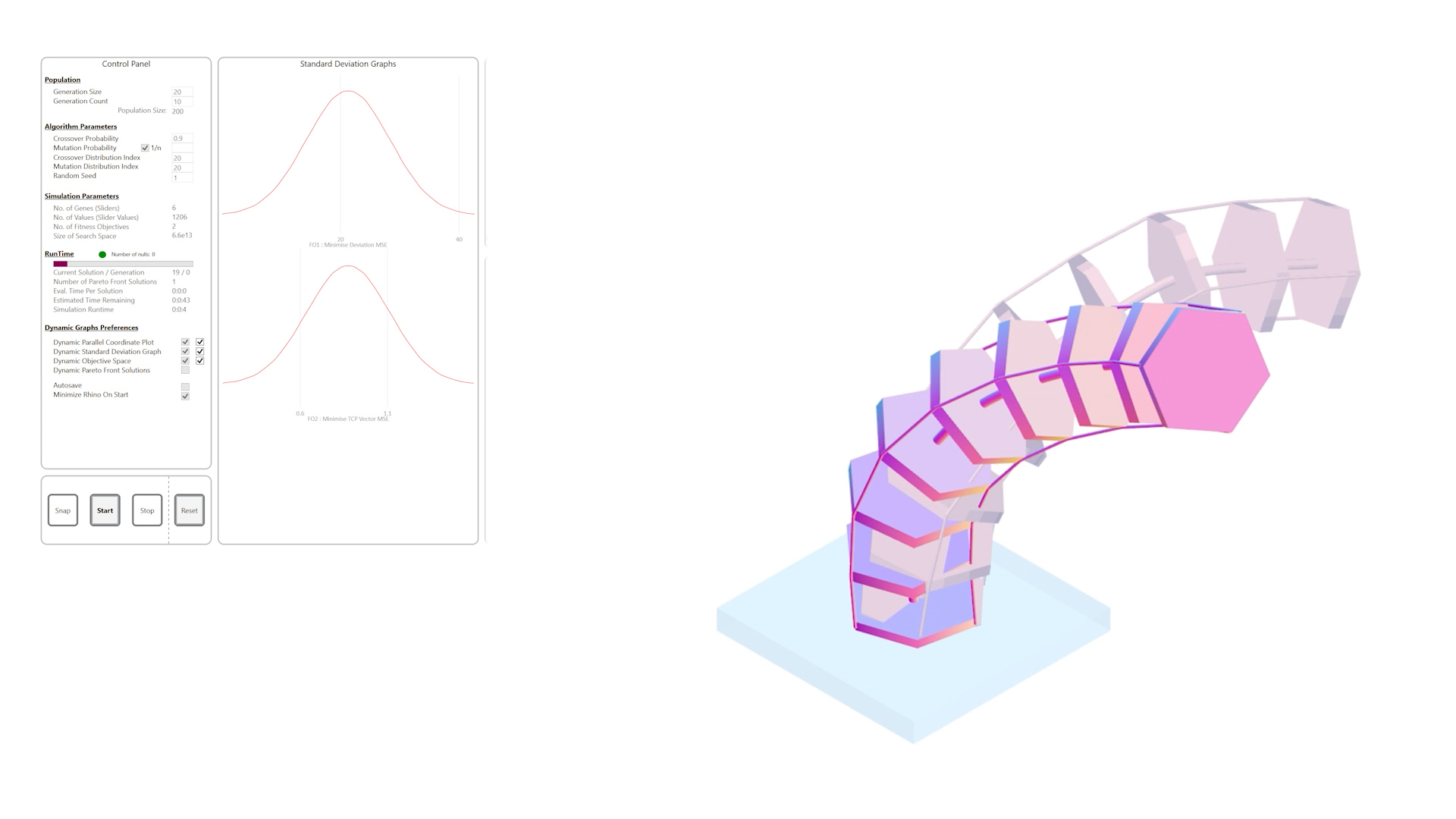}
 		 \caption{$t_0$.}
  		\label{fig:evo_1}
    \end{minipage}
    \hfill
    \begin{minipage}[t]{.3\textwidth}
        \centering
        \includegraphics[width=\linewidth]{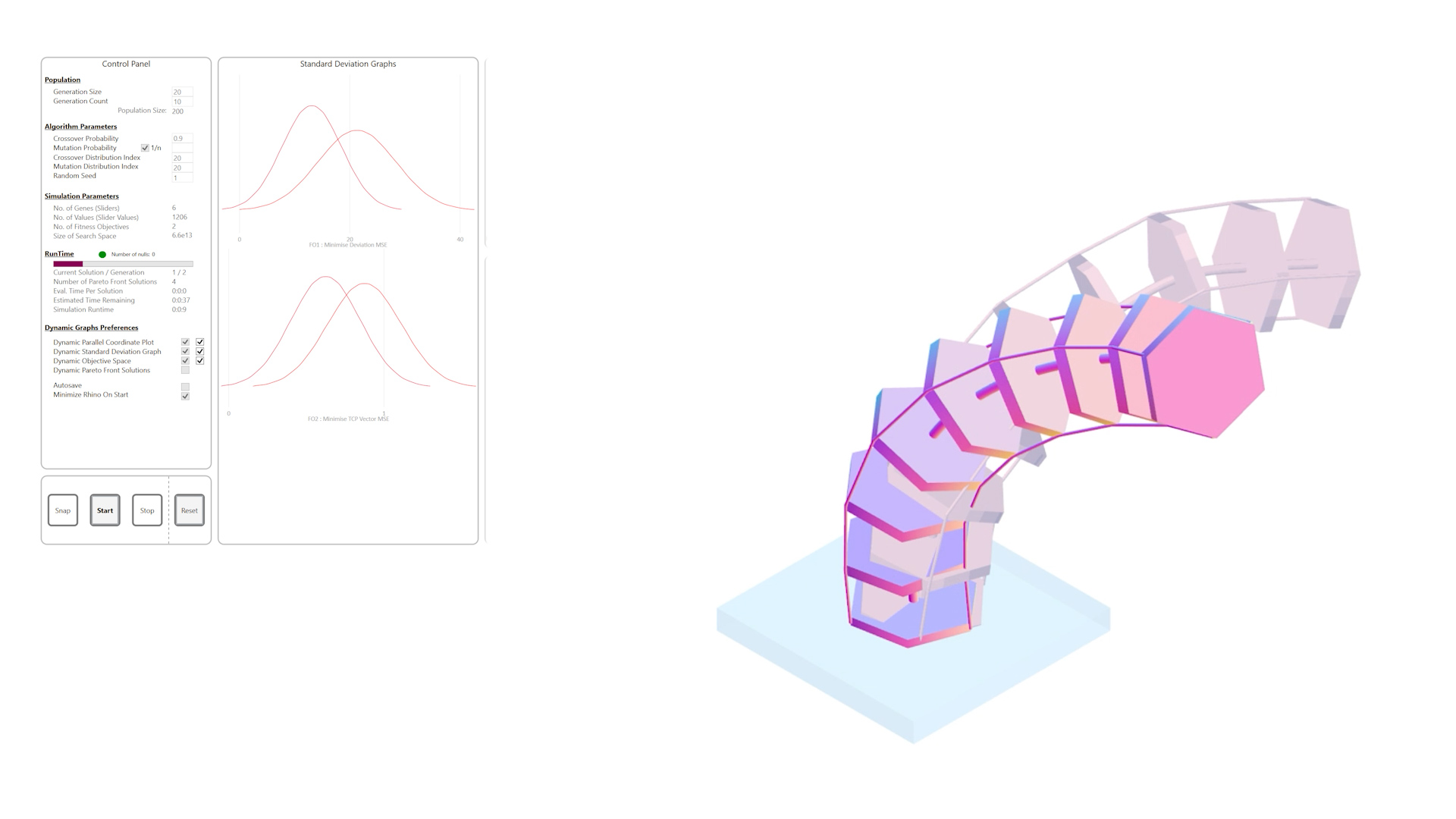}
 		 \caption{$t_1$.}
  		\label{fig:evo_2}
    \end{minipage}
	\hfill
    \begin{minipage}[t]{.3\textwidth}
        \centering
        \includegraphics[width=\linewidth]{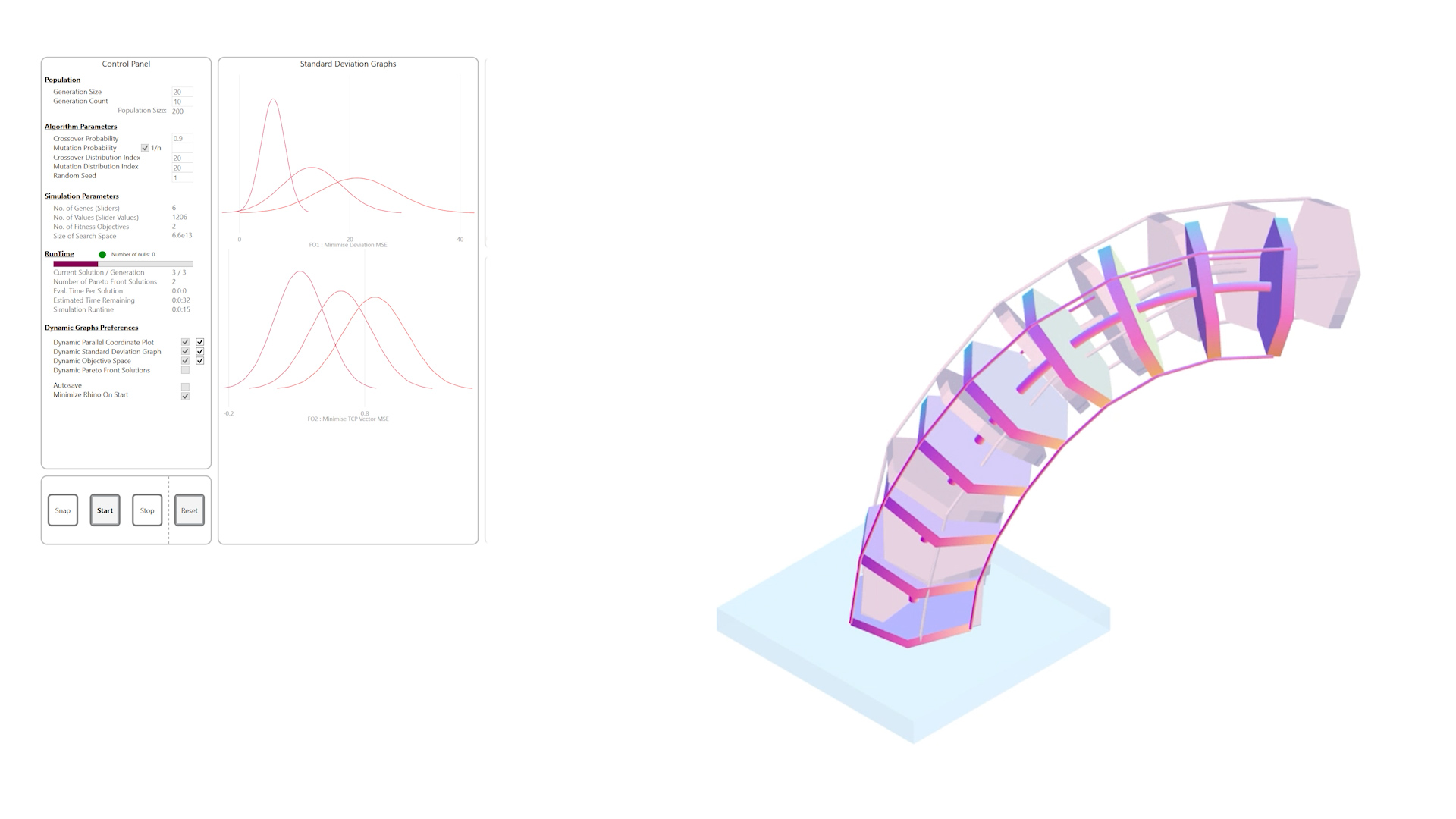}
 		 \caption{$t_2$.}
  		\label{fig:evo_3}
    \end{minipage}
\end{figure}

As for the digital modeling of continuum robot, figure \ref{fig:evo_1}, \ref{fig:evo_2}, and \ref{fig:evo_3} illustrate the optimization process over three distinct stages during the evolutionary algorithm, where the models in light color are the ideal configurations and the darker ones represent the models to be optimized. These stages demonstrate how the continuum robot's shape evolves over time, gradually approaching the ideal configuration by refining the coefficients through iterative optimization.

\begin{itemize}
	\item Initial Stage ($t_0$): In this stage, the backbone shape exhibits significant deviations from the ideal curve. This is due to the initial random coefficients used for the modal shape function, which result in a model that is far from optimized. At this point, the fitness objectives (\(\text{MSE}_1\) and \(\text{MSE}_2\)) are relatively high, indicating a large error.
	\item Intermediate Stage ($t_1$): As the evolutionary algorithm progresses through the first few generations, the robot configuration begins to show improvements. The coefficients are gradually refined, and the backbone curve aligns more closely with the ideal shape. At this stage, the fitness values start to decrease, reflecting the algorithm's ability to reduce both positional and TCP vector errors.
	\item Final Stage ($t_2$): In the final stage, the robot configuration closely matches the ideal shape. The backbone curve exhibits smooth bending patterns, with minimal deviations between the generated and target models. The evolutionary algorithm has effectively optimized the coefficients, resulting in convergence toward an optimal solution with low \(\text{MSE}_1\) and \(\text{MSE}_2\) values.
\end{itemize}

\begin{figure}[h]
	\centering
	\includegraphics[width=.85\linewidth]{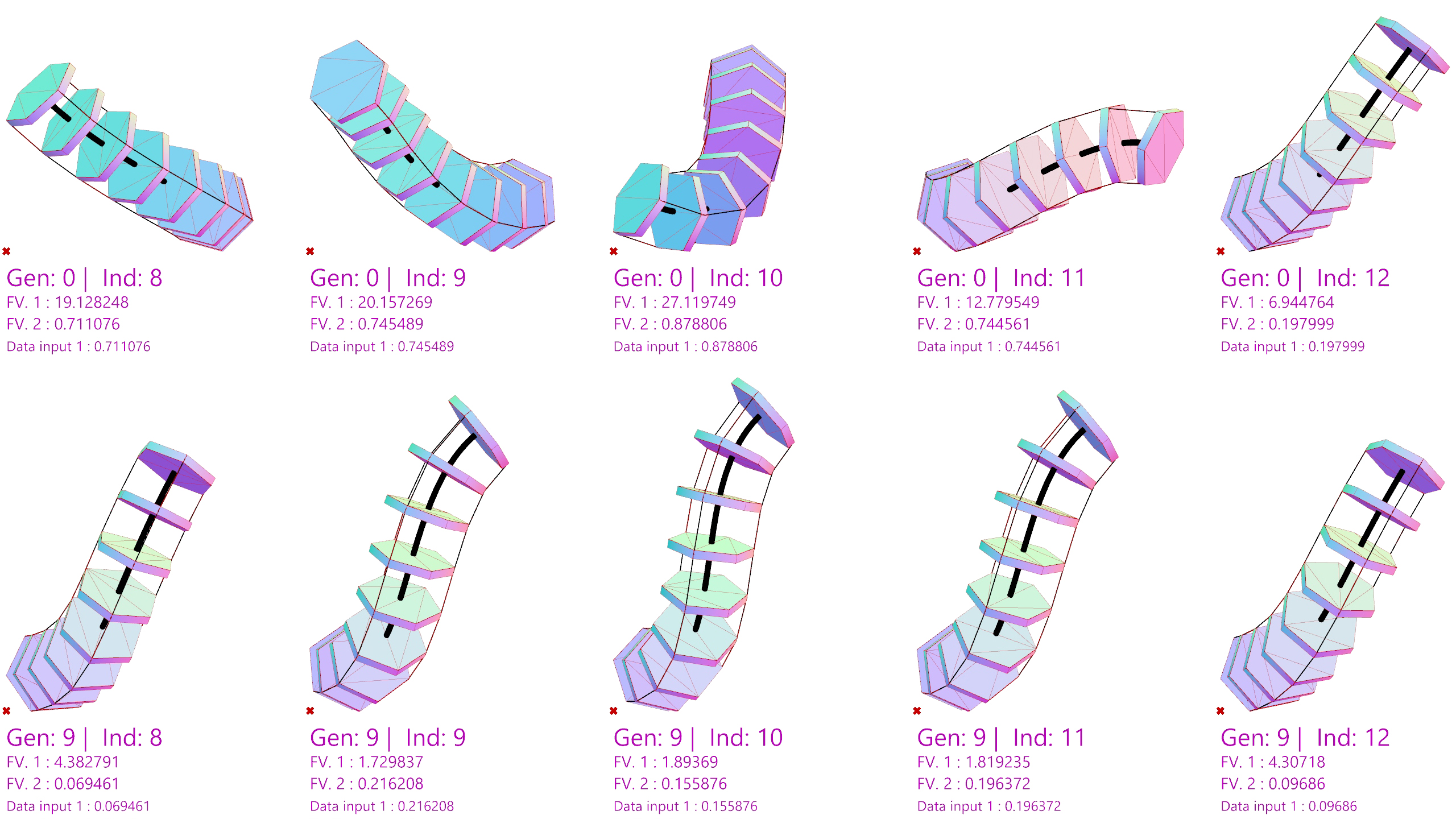}
	\caption{Modal shape derived from individual generation.}
	\label{fig:pheno_combine}
\end{figure}

\par The comparison between Generation 0 (the first generation) and Generation 9 (the last generation) highlights the effectiveness of the evolutionary algorithm in identifying the modal coefficients, with indexes from 8 to 12 randomly chosen (Figure \ref{fig:pheno_combine}). The individuals from Generation 0 (the first row) have a lower fitness value (FV), demonstrating less similarity to the ideal configuration. The individuals from Generation 9 (the second row), on the other hand, have been refined by the evolutionary algorithm with an improved FV.

\par The findings suggest that our method effectively reduces the computational complexity of identifying modal coefficients while maintaining high accuracy. This opens up possibilities for users and developers to model and test complex configurations more intuitively and efficiently.

\section{Limitations}
Despite its effectiveness, several limitations was found. First, the method relies heavily on predefined fitness objectives, which may not capture all the nuanced behaviors required for certain real-world applications, such as environmental interaction or advanced human-robot collaboration. In addition, the algorithm's performance might vary based on the complexity of the workspace and the number of required generations, leading to suboptimal results under some conditions.

\par Another drawback is the lack of direct integration with sensor data for real-time feedback, which limits the algorithm's applicability for dynamic environments. In scenarios such as medical surgery or wearable robotics, the absence of such feedback can hinder precise control and adaptation.

\section{Potential Applications}
There are several exciting avenues for future application of this method. One potential direction is the integration of real-time sensor data into the evolutionary framework to improve adaptability during operation. This enhancement could enable the deployment of continuum robots in dynamic environments, such as disaster response or surgical interventions.

\par Additionally, expanding the algorithm to support multi-objective optimization could allow for the consideration of more complex constraints, such as energy efficiency or material limitations. Another promising application is the adaptation of this framework for soft robots with environmental interactions, where deformation plays a crucial role in achieving desired behaviors.

\par Lastly, coupling the proposed framework with machine learning techniques could further refine the prediction of modal coefficients, enabling more intelligent control systems. By combining optimization with predictive models, future research could unlock new possibilities for autonomous, self-learning robotic systems.

\section{Conclusions}
This study presents a novel approach that integrates Lie group kinematics with evolutionary algorithms to address the inherent challenges of continuum robots in modeling and visualization. By automating the modal coefficient identification process within the CAD platform, our framework reduces the reliance on time-consuming physical experiments and manual parameter adjustments. This computational framework involves the digital modeling setup using the Chebyshev polynomials as basis functions, deployment of EA solver, and parametric settings of fitness objectives, bridging the gap between traditional kinematics and physics-based simulations. It offers a flexible and scalable modeling pipeline with less computational complexity. The results indicate that the integration of Lie group kinematics and evolutionary algorithm not only converges to optimal solutions efficiently but also ensures that the generated configurations align closely with the ideal robot models.

\par Overall, this research provides an alternative to modeling and visualization of continuum robots by integrating kinematics, optimization techniques, and computational design framework, aiming to enhance the accuracy and efficiency of complex simulations.

\bibliographystyle{unsrtnat}
\bibliography{references}  






\end{document}